# Research on an Autonomous UAV Search and Rescue System Based on the Improved EGO-Planner Algorithm

Haobin Chen
School of Information Science and Technology
Donghua University
Shanghai, China
hb_cccc@163.com

Junyu Tao
School of Information Science and Technology
Donghua University
Shanghai, China
ty579730@163.com

Bize Zhou
School of Mechanical Engineering
Donghua University
Shanghai, China
hf26023@163.com

Xiaoyan Liu*
School of Information Science and Technology
Donghua University
Shanghai, China
liuxy@dhu.edu.cn

*Abstract*—The demand is to solve the issue of UAV (unmanned aerial vehicle) operating autonomously and implementing practical functions such as search and rescue in complex unknown environments. This paper proposes an autonomous search and rescue UAV system based on an EGO-Planner algorithm, which is improved by innovative UAV body application and takes the methods of inverse motor backstepping to enhance the overall flight efficiency of the UAV and miniaturization of the whole machine. At the same time, the system introduced the EGO-Planner planning tool, which is optimized by a bidirectional A* algorithm along with an object detection algorithm. It solves the issue of intelligent obstacle avoidance and search and rescue. Through the simulation and field verification work, and compared with traditional algorithms, this method shows more efficiency and reliability in the task. In addition, due to the existing algorithm's improved robustness, this application shows good prospection.

## I. INTRODUCTION

In recent years, UAV technology has made a significant rise with the speed development of UAV sensors and flight control systems. Meanwhile, with the continuous update of Path-Planning technology, programming algorithms such as EGO-Planner that can be deployed in the UAV have emerged; thus, the autonomy and intelligence of UAV are improved, and it can be operated in a complex environment. UAVs have been widely used in logistics, security monitoring, and other hot fields. For example, the application of workshop patrol UAVs and automation-guided vehicles has proved the safety and efficiency that can be achieved by applying UAVs instead of humans in complex environments. The combined development of UAV and path-planning technology has drawn a broad blueprint for deploying UAVs in various fields [1].

This project works from three aspects: the improvement of the UAV's mechanical structure, the optimization of the path-planning algorithm, and verification training and application of the image recognition model. First, this project changes the mechanical structure of the UAV and then tries to deploy the path planning algorithm in real aircraft with image recognition technology. In terms of machinery, compared with the traditional UAV power mounting pull-up method [2][3], this paper improves the maneuverability and endurance of the UAV through the inverted motor design, and this design can install the payload above the propeller, making the whole machine more miniaturized and improving the overall flight efficiency. This project chooses RT-DETR as the object detection model for detecting trapped people. RT-DETR object detection model uses global self-attention to help the model capture complex relationships between objects and learn directly from input-to-output mapping without manually designed post-processing steps [4]. However, the disadvantages are high training costs and poor performance on small targets, which can be improved by data enhancement, multi-scale training, and post-processing strategies. The emerging EGO-Planner algorithm is selected as the planner first, and the bidirectional A* algorithm is used to optimize the planner through simulation [5]. Finally, the excellent performance of the improved algorithm is verified in the on-site tests.

This paper proposes an autonomous search and rescue UAV system based on an EGO-Planner algorithm that is improved by an innovative UAV body application. It improves the UAV's intelligence and overall flight performance. Meanwhile, simulation and field verification show that the robustness of this method is greatly improved compared with the existing methods so that it can better complete its application in complex environments.

## II. THE UVA SYSTEM

Miniaturization design of the UAV fuselage (see figure 1), with the only wheelbase of 206.46mm, and the design of an inverted motor, that is, the propeller is under the arm to reduce the interference of the arm on the airflow. Compared with the traditional configuration (see Figure 2) [2][3], this design has greater thrust output [6], which improves the overall flight efficiency of the UAV. At the same time, sensors, onboard computers, and other payloads can also be installed above the propeller, thus making the fuselage miniaturized while still capable of enough maneuverability and endurance. The electric regulator and power module are installed in the center of the four rotors. When the propeller rotates, a large amount of gas will flow through it to dissipate heat, therefore improving the power stability of the entire system. The tripod is installed at the bottom and is flexibly connected to the fuselage by silicone pads to reduce the impact on the electronics of the UAV during landing. The tripod shape is designed to avoid the airflow generated by the blades on a vertical projection(see Figure 3).

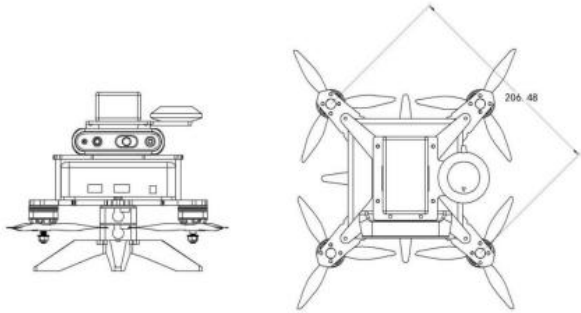

Fig. 1. Front view (the left figure) Bird's eye view (the right figure)

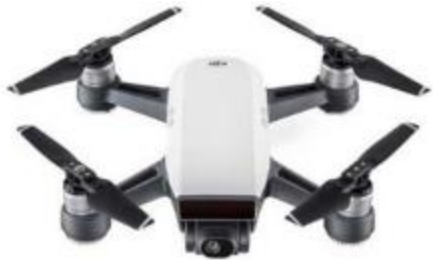

Fig. 2. Multi-rotor UAV in the traditional configuration

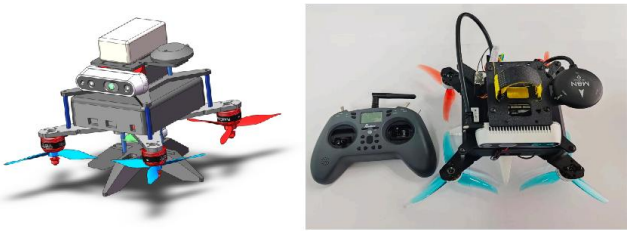

Fig. 3. 3D model and Object picture

The hardware composition of UAV:

The hardware of the UAV system includes a flight power system, a primary flight control system, and a perception and planning system. Hardware modules are shown in Figure 4.

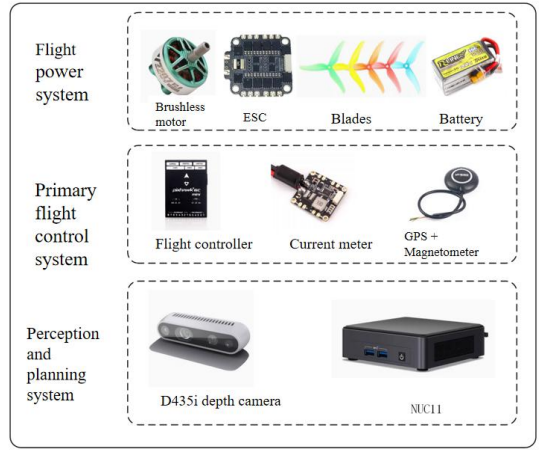

Fig. 4. UAV hardware diagram

The flight power system provides the thrust and power needed to maneuver the UAV. The flight power system consists of a motor, electric regulator, blade, battery, current meter, and voltage regulation module.

The primary flight control system is responsible for the attitude control of UAVs, and it is the lowest and most basic control system of UAVs. It consists of Pixhawk6c mini, GPS, and magnetometer, as shown in Table 1.

The perception and planning system is responsible for real-time perception of the environment, acquisition of point cloud data and RGB images, hull localization and target acquisition, and autonomous trajectory planning, which is critical for the UAV's autonomous search and rescue tasks. The system consists of hardware such as a D435i depth camera and NUC11 onboard computer, as well as algorithm software.

TABLE I. UAV HARDWARE PARAMETERS

| Hardware type | Hardware model |
|---|---|
| Flight controller | Pixhawk 6cmini |
| Motor | T-Motor V2207 KV1950 |
| Battery | 6s 1500mAh |
| Depth camera | D435i |
| Airborne computer | NUC11 |

### III. IMPROVED EGO-PLANNER ALGORITHM

For the motion planning of agent motion, the traditional method generally adopts a gradient-based planner; that is, gradient magnitude and direction are evaluated using a pre-built ESDF map [7]. Numerical optimization is used to generate the local optimal solution. However, it is tough to establish the ESDF diagram, and the calculation of ESDF accounts for 70% of the total local planning time. Therefore, the construction of ESDF has become the bottleneck of gradient-based planners.

Currently, the methods of constructing ESDF maps are divided into incremental full update and batch local calculation, namely Voxblox and FIESTA. However, neither method pays attention to the trajectory. Therefore, too much calculation is spent on ESDF values that do not contribute to planning. In other words, the current ESDF-based methods cannot directly serve trajectory optimization.

Therefore, based on the traditional method, we use EGO-Planner (ESDF-free Gradient-based lOcal planning) [8], a local path planning framework without ESDF as path navigation. It consists of a gradient-based spline optimizer and a post-refinement program. Compared with traditional methods, it is more lightweight and intelligent. The following is a detailed description of the construction process of EGO-Planner.

*A. Path Planning*

First, it is necessary to generate a collision-free and kinematics-constrained initial path through path planning and use it as the initial path. It will significantly improve the success rate and efficiency of subsequent planning and optimization, reduce the consumption of computing resources, and improve the model's performance. Time required for various pathfinding algorithms

A* algorithm is usually used in the trajectory planning of the classic EGO-Planner algorithm. Through theoretical derivation and practice, this study has demonstrated that under the same computing power, the bidirectional A* algorithm (see Figure 6) can be used to replace the original A* algorithm in the planner (see Figure 5). At the same time, the search time can be reduced by up to 50% [9]while ensuring the optimal path [10](see Table 2).Bidirectional A* algorithm flow chart is shown in Figure 7.

TABLE II. TIME REQUIRE FOR VARIOUS PATHFIND ALOGRITHMS

| algorithm | bidirectional A* | kinodynamic A* | Dijkstra |
|---|---|---|---|
| Time | 1.5s | 2.3s | 9s |
| Speed | 2m/s | 1.34m/s | 0.34m/s |

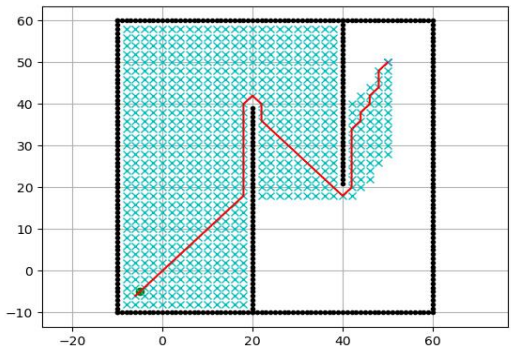

Fig. 5. The schematic diagram of the classical A* algorithm

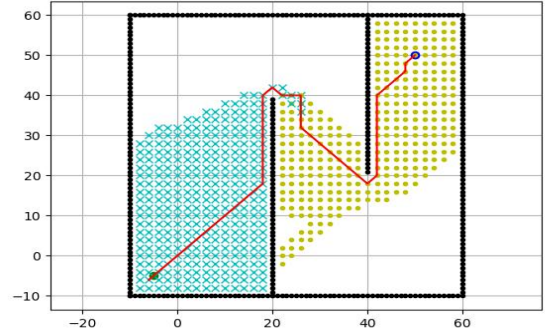

Fig. 6. The schematic diagram of the bidirectional A* algorithm

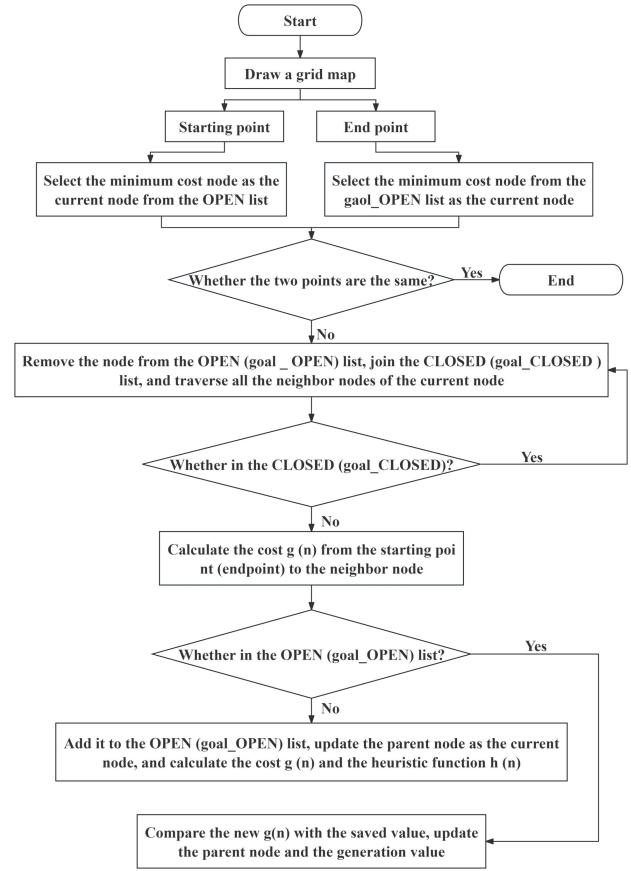

Fig. 7. Bidirectional A* algorithm flow chart

*B. Collision Avoidance Force Calculation*

Based on the collision-free trajectory generated in the first step, a B-spline curve $\Phi$ is generated by using the B-spline curve method [11], which meets the constraints conditions but does not consider obstacles. For each collision segment detected during the iteration, a collision-free path $\Gamma$ is generated by the above bidirectional A* algorithm, and the plane $\Psi$ perpendicular to the tangent vector $R_i$ intersects with

Γ to form a line. Determine the p, v pair from it. For each control point $Q_i$ of the collision line segment, a positioning point $p_{ij}$ is generated on the obstacle surface, and a corresponding repulsion direction vector $v_{ij}$ is in the same direction as $Q_i p_{ij}$.

IV. TRAJECTORY OPTIMIZER BASED ON GRADIENT

A. Modeling

Based on the convex-hull property and derivative properties of uniform b-spline curves, we can obtain the first-order, second-order, and third-order derivative (V, A, J) control points of the trajectory Φ.

$$V_i = \frac{Q_{i+1} - Q_i}{\Delta t}, A_i = \frac{V_{i+1} - V_i}{\Delta t}, J_i = \frac{A_{i+1} - A_i}{\Delta t} \quad (1)$$

Therefore, according to the differential flatness characteristics of the UAV [12], it is only necessary to control the x, y position, and yaw angle Ψ of the UAV. All the remaining states can be expressed by the algebraic combination of these three variables and their finite-order derivatives, thereby reducing the amount of control we need to plan. Based on the above, the optimization problem is redefined as.

$$\min_Q J = \lambda_s J_s + \lambda_c J_c + \lambda_d J_d \quad (2)$$

Among them, $J_s$ is the smoothing term penalty, $J_c$ is the collision term penalty, $J_d$ is the feasibility penalty, and $\lambda$ is the penalty term weighted value.

1) Smoothing Term Penalty

In B-spline trajectory optimization, the smoothing term penalty is formulated as a time integral on the trajectory parameters (a, J, etc.) and the square derivative. Because of the convex hull property of B spline, the (A and J) sum of squares of acceleration and acceleration change rate can be effectively reduced by minimizing the sum of squares of the second and third order control points of Φ. The formula is as follows.

$$J_s = \sum_{i=1}^{N_c-2} \|A_i\|_2^2 + \sum_{i=1}^{N_c-3} \|J_i\|_2^2 \quad (3)$$

2) Collision Term Penalty

Collision penalty keeps control points away from obstacles. It is realized by using safety gap and penalty control points. To further optimization, a quadratic continuous differentiable penalty function is constructed. Furthermore, with the decrease of the $d_{ij}$ inhibition of its slope, a piecewise function is obtained.

$$j_c(i,j) = \begin{cases} 0, & c_{ij} \leq 0 \\ c_{ij}^3, & 0 \leq c_{ij} \leq s_f \\ 3s_f c_{ij}^2 - 3s_f^2 c_{ij} + s_f^3, & c_{ij} > s_f \end{cases} \quad (4)$$

$$c_{ij} = s_f - d_{ij} \quad (5)$$

The total collision term penalty is obtained by summing the penalties of all control points.

$$J_c = \sum_{i=1}^{N_c} j_c(Q_i) \quad (6)$$

The gradient is obtained by directly calculating the derivative of the quadratic continuous differentiable penalty function to $Q_i$.

$$\frac{\partial j_c}{\partial Q_i} = \sum_{i=1}^{N_c} \sum_{j=1}^{N_P} \begin{cases} 0, & c_{ij} \leq 0 \\ -3c_{ij}^2, & 0 \leq c_{ij} \leq s_f \\ -6s_f c_{ij} + 3s_f^2, & c_{ij} > s_f \end{cases} \quad (7)$$

3) Feasibility Term Penalty

By limiting the high-order derivatives of the trajectory in each dimension, its feasibility is guaranteed. Due to the convex hull property of the B-spline, the derivative constraint on the control points is sufficient to constrain the entire B-spline curve. Let F (·) be the penalty function constructed for the higher-order derivatives of each dimension.

$$f(c_r) = \begin{cases} a_1 c_r^2 + b_1 c_r + c_1, & c_r < -c_j \\ (-\lambda c_m - c_r)^3, & -c_j < c_r < -\lambda c_m \\ 0, & -\lambda c_m \leq c_r \leq \lambda c_m \\ (c_r - \lambda c_m)^3, & \lambda c_m < c_r < c_j \\ a_2 c_r^2 + b_2 c_r + c_2, & c_r > c_j \end{cases} \quad (8)$$

$$F(C) = \sum_{r=x,y,z} f(c_r) \quad (9)$$

B. Numerical Optimization

For the objective function:

$$\min_Q J = \lambda_s J_s + \lambda_c J_c + \lambda_d J_d \quad (10)$$

With the continuous addition of new obstacles will continue to change, which requires the solver to be able to quickly restart and solve. In addition, the objective function is mainly composed of quadratic terms, and the use of Hessian matrix information can speed up the convergence rate. However, obtaining an accurate H matrix will consume a lot of computer resources. Therefore, we use the quasi-Newton method to approximate the H matrix from the gradient information.

Moreover, further optimization of time allocation and trajectory are as follows. Based on the safe trajectory $\Phi_s$ obtained in the previous step, a uniform B-spline curve $\Phi_f$ with reasonable time allocation is re-generated after reasonable time allocation. Therefore, using the method of anisotropic curve fitting, $\Phi_f$ is allowed to freely optimize its control points while maintaining the derivative shape which is almost the same as that of $\Phi_s$. So that it could meet the higher-order derivative constraints.

The optimization steps are as follows.

Firstly, Calculate the limit exceeding rate. (The maximum proportion exceeding the limit. The subscript m denotes the maximum value of the restriction.)

$$r_c = \max\{|V_{i,r}/v_m|, \sqrt{|A_{j,r}/a_m|}, \sqrt[3]{|J_{k,r}/j_m|}, 1\} \quad (11)$$

$r_c$ indicates how much time $\Phi_f$ needs to allocate for $\Phi_s$.

Secondly, obtain new time interval of $\Phi_f$ is obtained.

$$\Delta t' = r_c \Delta t$$

Thirdly, the new time interval is used to generate trajectory $\Phi_f$, while maintaining the same shape and control points as $\Phi_s$. Furthermore, the smoothing term penalty and the feasibility term penalty are recalculated to obtain a new objective function:

$$\min_Q J' = \lambda_s J_s + \lambda_d J_d + \lambda_f J_f \quad (12)$$

Finally, $\lambda_f$ is the fitting term weight. $J_f$ is defined as an anisotropic displacement integral from $\Phi_f(\alpha T')$ to $\Phi_s(\alpha T)$ ($\alpha \in [0,1]$). $T$ and $T'$ are the duration of the trajectory $\Phi_s$ and $\Phi_f$. Based on the above, for $\Phi_f$, the smooth adjustment limit is relaxed by using the low-weight axial displacement $d_a$. Highly weighted radial displacement $d_r$ is used to prevent collisions.

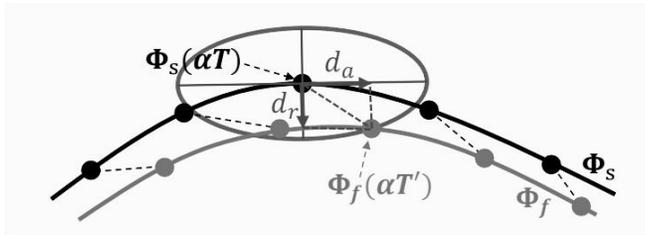

Fig. 8. Displacement diagram of diameter (shaft)

The axial displacement is the tangent direction of the point. The radial displacement is the vertical direction of the tangent of the point (in Figure 8). The final effect of trajectory optimization is shown in Figure 9.

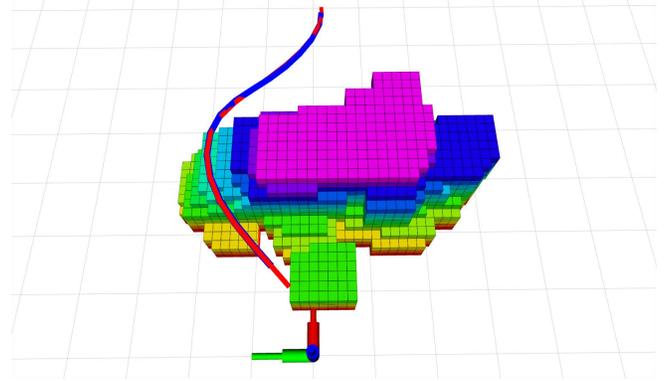

Fig. 9. Simulation planning effect diagram

## V. SEARCH AND RESCUE IDENTIFICATION BASED ON RT-DETR TARGET DETECTOR

### A. Introduction of RT-DETR

RT-DETR（Real-Time DEtection TRansformer）[13], a new type of real-time end-to-end target detector proposed by Baidu researchers. This work implements an end-to-end detector that does not rely on post-processing (such as non-maximum suppression NMS) in real-time target detection tasks for the first time (see figure 10).

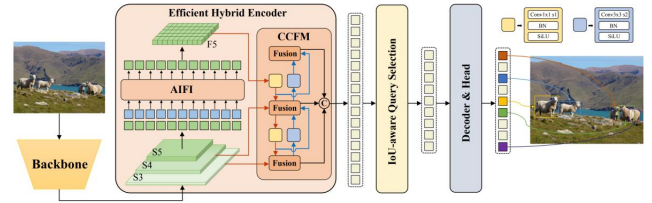

Fig. 10. Overview of RT-DETR

Depending on the depth and width of the network, RT-DETR offers several versions in different sizes. The introduction of multi-scale features is beneficial to accelerate training convergence and improve performance.[14] For the UAV platform, considering the weak performance of the computer, the RT-DETR-L model is selected. RT-DETR-L achieves an AP (average precision) of 53.0% and a performance of 3 FPS (frames per second) on the COCO val2017 dataset.

### B. Model Training

First, the experimenter prepares the data set, simulates the posture of the trapped person to take 1000 photos at different angles, backgrounds, and light, and uses Labellmg software to mark the pictures. In addition, 1000 photo samples were randomly divided into 700 training sets, 200 validation sets, and 100 test sets. Then the PaddlePaddle deep learning framework [15] is called to train RT-DETR. The model hyperparameters are shown in Table 3.

TABLE III. TRAINING HYPERPARAMETERS

| Parameter | Settings | Parameter | Settings |
|---|---|---|---|
| epochs | 400 | weight_decay | 0.0001 |
| batch | 16 | milestones | [100,200,300] |
| base_lr | 0.0001 | use_warmup | true |
| optimizer | AdamW | clip_grad_by_norm | 0.1 |
| gamma | 0.1 | workers | 4 |

*C. Result Display*

The trained weight model is used to identify, and the recognition results are shown in Figure 11.

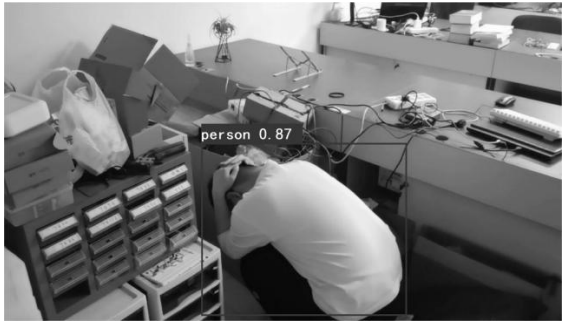

Fig. 11. Recognition results

## VI. CONCLUSION

This paper proposes a UAV system based on the EGO-Planner algorithm optimized by UAV applications, which has an improved structure to perform autonomous search and rescue tasks in various complex environments. The practical results show that the improved system has many advantages compared with the traditional system, including high flight efficiency, high autonomy, strong recognition ability, and robustness.

However, the endurance of the UAV autonomous control system in this study is greatly affected by the weather, the training time required for target detection is long, and the adaptability to different targets is low, so improvements are needed. In the future, for the application and optimization of UAV swarm control technology, we need to adopt a more suitable model to complete the detection of diversified targets. In addition, the UAV body structure will be further improved to meet the endurance requirements.